\documentclass[runningheads]{llncs}
\usepackage[T1]{fontenc}
\usepackage{amsmath}
\usepackage{amssymb}    
\usepackage{graphicx,verbatim}
\usepackage{xcolor}
\usepackage{hyperref}
\usepackage{mathtools}
\usepackage{soul}
\begin{document}
%
\title{When Confidence Lacks Concepts: Interpretable OOD Detection via Representation Perturbations}

%

\author{Anju Chhetri$^{1}$, 
        Pratik Shrestha$^{1}$,
        Ramesh Rana$^{3}$,
        Sam Philip$^{5}$,
        Prashnna Gyawali$^{2}$,
        Binod Bhattarai$^{1,4,6}$}  
\authorrunning{A. Chhetri et al.}
\institute{$^{1}$NepAl Applied Mathematics and Informatics Institute for research, Nepal\\
$^{2}$West Virginia University, USA\\
$^{3}$Kathmandu University, Nepal\\
$^{4}$University College London, UK\\
$^{5}$NHS Grampian, UK\\
$^{6}$University of Aberdeen, UK\\
    }
  
\maketitle              
\begin{abstract}
Deep neural networks have achieved remarkable performance across medical imaging tasks, yet their tendency to overgeneralize under distributional shifts poses a major obstacle to safe clinical deployment.
Out-of-Distribution (OOD) detection methods aim to mitigate this risk, but most existing approaches rely on opaque internal signals with poorly understood semantic meaning, limiting trust in safety-critical settings.
In this work, we propose an interpretable OOD detection framework that probes the stability of model predictions under class-conditioned semantic perturbations.
Leveraging sparse autoencoders (SAEs), we learn class-specific concept vectors from in-distribution data that disentangle dense intermediate representations into sparse, semantically meaningful components.
At inference, we perturb deeper-layer representations using the concept vectors associated with the model’s predicted class and measure the 
class logits stability.
We hypothesize that in-distribution samples exhibit low sensitivity to such perturbations, as their representations align with class-specific semantic directions, whereas OOD samples show amplified deviations due to representational misalignment.
By framing OOD detection as a concept-conditioned stability analysis, our approach provides both a discriminative OOD signal and an interpretable lens into the internal mechanisms driving model uncertainty, making it particularly suitable for high-stakes medical applications.


\end{abstract}
\section{Introduction}\label{introduction}

Despite recent advances in deep learning across diverse domains \cite{krizhevsky2012imagenet,he2016deep}, these models suffer from overgeneralization, limiting real-world deployment where distributional shifts are inevitable \cite{taori2020measuring}. 
This issue is particularly critical in healthcare, where variations in imaging protocols, patient demographics, or the emergence of novel diseases may lead to severe diagnostic errors. To mitigate these risks, Out-of-Distribution (OOD) detection has emerged as a principled framework for identifying inputs that deviate from the training distribution before unreliable predictions are made \cite{tschuchnig2021anomaly}.
OOD detection has received increasing attention, with most approaches proposing post-hoc scoring functions that distinguish in-distribution (ID) from OOD samples without requiring model retraining. Early methods operate directly on network outputs, such as Maximum Softmax Probability (MSP) baseline \cite{hendrycks2016baseline}, while subsequent refinements introduce energy-based scores \cite{liu2020energy} or maximum logit values \cite{hendrycks2019scaling}.

Beyond output-space signals, distance-based techniques measure deviations in feature space, for example, using Mahalanobis distance to class-conditional distributions \cite{lee2018simple}. Feature manipulation strategies further enhance separability between ID and OOD samples, including approaches such as ODIN \cite{liang2017enhancing}, I-ODIN \cite{regmi2024image}, ReAct \cite{sun2021react}, and NECO \cite{ammar2023neco}. Hybrid methods, such as ViM \cite{wang2022vim}, combine feature and logit information, while gradient-based techniques like GradNorm \cite{huang2021importance} leverage training dynamics for OOD scoring.
More recently, relevance-based approaches, NERO \cite{chhetri2025nero}, exploit relevance distances as discriminative signals.

Despite this methodological diversity and steady performance gains, most OOD detection methods rely on internal model signals 
whose semantic meaning remains opaque. Although these approaches can successfully identify anomalous inputs, they offer limited insight into the features or reasoning underlying the detection decision. This limitation reflects a broader black box challenge in modern machine learning, where increasingly large and highly performant architectures, including Vision Transformers (ViT) \cite{dosovitskiy2020image}, achieve state-of-the-art results while their internal decision making processes become progressively less transparent \cite{elhage2022toy}. As model capacity continues to scale, the interpretability gap widens, raising concerns about trust, accountability, and reliability in safety critical domains such as healthcare.
In such settings, effective OOD detection must therefore provide not only accurate anomaly identification but also remain interpretable. Recent approaches have explored interpretability in OOD detection through concept-based explanations and relevance-based attribution methods, providing insights into the factors underlying detection decisions \cite{choi2023concept,chhetri2025nero}.

We address this challenge by leveraging recent advances in understanding the representational underpinnings of deep neural networks.
Deep networks are known to organize ID samples into structured, class-conditional regions of representation space, with deeper layers exhibiting increasingly semantically meaningful clustering \cite{ren2024deep,raghu2017svcca}.
Under distribution shift, inputs may deviate from these learned regions, resulting in representational misalignment and unstable predictions \cite{tamang2025handling,morningstar2021density}.
Building on these insights, we introduce a framework to assess logit stability via class-conditioned perturbations in the representation space.
We utilize Sparse Autoencoders (SAEs) to decompose dense activations into a basis of interpretable concept vectors \cite{arad2025saes}, which serve as the linear directions encoding class-specific features. 
Guided by the linear representation hypothesis \cite{park2023linear}, we then quantify the stability of the model's prediction by performing controlled perturbations along these learned directions and measuring the resulting change in class logits, thereby assessing how strongly a prediction relies on the underlying class-defining features.
We demonstrate that ID samples, already aligned with class-relevant semantic directions (empirically validated in Fig.~\ref{fig:robustness}(a)), exhibit low sensitivity to these perturbations, producing small logit changes (Lemma~\ref{lem:ln_sensitivity}), whereas OOD samples show larger deviations due to misalignment with the learned class-conditioned directions. Our contributions are threefold:

    \begin{enumerate}
        \item We propose \underline{C}lass-conditioned \underline{A}ctivation \underline{P}erturbation\underline{S} (CAPS), a novel OOD detection framework that leverages concept-level perturbations in representation space.
        \item We extensively evaluate our approach on multiple medical imaging domains, including endoscopy, histopathology, and 
        ophthalmic imaging, demonstrating its effectiveness across anatomically and visually distinct modalities.
        \item We demonstrate that our extracted features correspond to clinically meaningful visual patterns through clinical expert validation and subsequently utilize these patterns as a basis for robust OOD detection.
    \end{enumerate}

\section{Method}\label{method}
\begin{figure}[!b] 
    \centering
    \includegraphics[trim= 0cm  0cm 0cm 0cm, width=1\textwidth]{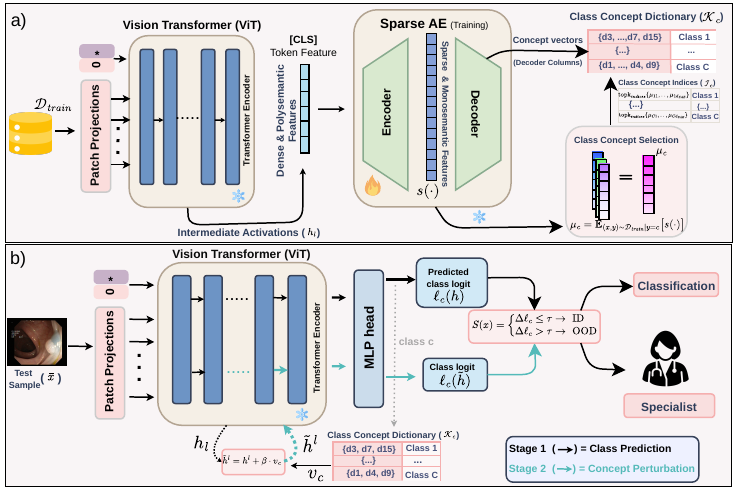} 
\caption{
\textbf{(a) SAE Training:} Intermediate ViT activations are used to train an SAE; sparse coefficients are aggregated per class to select top-$k$ decoder columns, forming a class concept dictionary. 
\textbf{(b) Inference:} For a test sample $\bar{x}$, we first obtain the predicted class from the base model and select the corresponding concept set. These concepts are aggregated into a class-specific concept vector $v_c$, which is injected into the intermediate activation. The resulting change in the predicted-class logit ($\Delta \ell_c$) is then used as the OOD detection score.
}
    \label{fig:method}
\end{figure}
\noindent\textbf{Problem Formulation}: We consider a multi-class classification task with training and testing sets 
    $\mathcal{D}_{\text{train}} = \{(x_i, y_i)\}_{i=1}^{n_t}$ and 
    $\mathcal{D}_{\text{test}} = \{(x_i, y_i)\}_{i=1}^{n_e}$, 
    where $x$ is an image and $y \in \{0,\dots,C-1\}$ its class label, with $C$ total classes. 
    We assume the data are independent and identically distributed (i.i.d.) from $\mathcal{P}_{\text{id}}$. 
    Let $f_w: \mathcal{X} \to \mathbb{R}^C$ be a neural network trained, hereafter referred to as the base model, on $\mathcal{D}_{\text{train}}$, parameterized by $w$, and denote the class-token activation at layer $l$ as $f_w^l(x) \in \mathbb{R}^{d_{\text{model}}}$. 
    The goal of OOD detection is to derive a confidence score that identifies whether a test sample is drawn from $\mathcal{P}_{\text{id}}$ or from an unknown OOD distribution $\mathcal{P}_{\text{ood}}$.\\
\noindent\textbf{Training SAEs:}  
Sparse autoencoders (SAEs), initially applied to study representations in large language models, have recently been shown to be effective for vision models as well \cite{joseph2025prisma,stevens2025interpretabletestablevisionfeatures}.  
Here, we employ SAEs to model intermediate class-token [CLS] activations of a ViT.  
Given the residual-stream activation at layer $l$, $h \in \mathbb{R}^{d_{\text{model}}}$ from the base model, the SAE learns a sparse decomposition:
\begin{equation}\label{eq:1}
h \approx b + \sum_{i=1}^{d_{\text{SAE}}} s_i(h) d_i,
\end{equation}
where $b$ is a learned bias, $s(h)\in\mathbb{R}^{d_{\text{SAE}}}$ is the sparse latent representation, and $\{d_i\}$ are columns of the learned decoder $D$.  
The encoder $s: \mathbb{R}^{d_{\text{model}}} \to \mathbb{R}^{d_{\text{SAE}}}$ maps dense activations to a higher-dimensional sparse space, while the decoder reconstructs the original activation, denoted by $\bar{h}$, via a sparse linear combination of features.  
The SAE is trained to balance reconstruction fidelity and sparsity,
 \begin{equation}\label{eq:2}
 \mathcal{L} = \|h - \bar{h}\|_2^2 + \gamma \|s(h)\|_1,
 \end{equation}
\noindent\textbf{Collecting Class-Specific Concepts:}
\label{method:collectingConcept}
For each class $c$, only a subset of SAE latents are characteristically active \cite{stevens2025interpretabletestablevisionfeatures}. We identify these by computing the mean latent activation over training samples of class $c$:
\begin{equation}\label{eq:3}
\mu^c = \mathbb{E}_{(x,y)\sim \mathcal{D}_{train} \mid y = c}\big[ s(f_w^{l}(x)) \big],
\end{equation}
We then select the top-$k$ latent dimensions with the largest values in $\mu^c$, denoted by $\mathcal{I}_{c}$.
These high-magnitude activations in SAEs correspond to latent directions most associated with class-discriminative
concept components \cite{arad2025saes}.
The corresponding decoder columns from the SAE decoder matrix $D$ define the class concept dictionary $\mathcal{K}_c = \{ D_i \mid i \in \mathcal{I}_c \}$. Unless otherwise stated, we set $k$=20.\\

\noindent\textbf{Class-Conditioned Activation Perturbation: }
\label{method:conceptbasedintervention}
At inference time, given a test input 
$\bar{x}$,
the network predicts a class
$\hat{y}$
$= \arg\max_{c}$
$f_w(\bar{x})$.
We then retrieve the corresponding concept set
$\mathcal{K}_{\hat{y}}$
and aggregate its elements by summation, followed by
$l_2$-normalization, to obtain a single class concept vector $v_{\hat{y}}$.
We then perform a class-conditioned activation perturbation at layer
$l$
by adding  $v_{\hat{y}}$ to the activation, multiplied by $\beta$, which controls the strength of the perturbation.
\begin{equation}\label{eq:6}
\tilde{h}^{l} = h^{l} + \beta v_{\hat{y}},
\end{equation}
The modified activation $\tilde{h}^l$ is propagated forward through the remaining layers of the network while keeping all parameters fixed.
We target the [CLS] token ($h\in \mathbb{R}^{d_{\text{model}}}$) as it provides a compact, global representation of the image features.
The choice of the residual stream aligns with established practices in mechanistic interpretability for performing linear interventions \cite{bereska2024mechanistic}.
Furthermore, we target the final layer because it exhibits the most semantically refined, class-specific features \cite{ren2024deep} (Sec.~\ref{introduction}) and enables closed-form sensitivity analysis.
\begin{lemma}[First-order logit sensitivity under concept perturbation]
\label{lem:ln_sensitivity}
Let $h \in \mathbb{R}^{d_{model}}$ denote the final-layer representation, and assume the classifier has the form
$\ell_c(h)=w_c^\top \phi(h)$, where $\phi:\mathbb{R}^{d_{model}}\!\to\!\mathbb{R}^{d_{model}}$.
For a concept direction $v_c$ and sufficiently small $\beta v_c>0$, first-order Taylor expansion around $h$ gives:
\begin{equation}
\Delta \ell_c
\coloneqq
\ell_c(h+\beta v_c)-\ell_c(h)
\;\approx\;
\beta\, w_c^\top J_\phi(h)\, v_c,
\label{eq:first_order_general}
\end{equation}
where $J_\phi(h)$ is the Jacobian of $\phi$ at $h$.
\end{lemma}

\noindent
\textit{ViT case.}
In Vision Transformers, $\phi(h)=\mathrm{LN}(h)$. Using the LayerNorm Jacobian \cite{xiong2020layer}, where ${d = d_{model}}$,
\[
J_{\mathrm{LN}}(h)=
\frac{1}{\sigma(h)}
\Big(
I-\frac{1}{d}\mathbf{1}\mathbf{1}^\top
-\frac{(h-\mu(h))(h-\mu(h))^\top}{d\,\sigma(h)^2}
\Big),
\]
Eq.~\eqref{eq:first_order_general} yields
\begin{equation}
\Delta \ell_c
\approx
\frac{\beta}{\sigma(h)}
\left[
w_c^\top v_c
-\frac{w_c^\top \mathbf{1}\mathbf{1}^\top v_c}{d}
-\frac{(h-\mu(h))^\top v_c}{d\,\sigma(h)^2}
\, w_c^\top (h-\mu(h))
\right].
\label{eq:first_order_explicit}
\end{equation}
Under Lemma~\ref{lem:ln_sensitivity}, we define the \emph{logit sensitivity score} as the predicted-class logit shift $\Delta \ell_c$ induced by a class-concept perturbation. 
In the LayerNorm case, the term $(h-\mu(h))^\top v_c$ captures the alignment between the concept direction and the centered representation of a test sample $\bar{x}$. 
As discussed in Sec.~\ref{introduction} and empirically illustrated in Fig.~\ref{fig:robustness}(a), this alignment is typically stronger for ID samples and weaker under distribution shift. 
We therefore use $\Delta \ell_c$ as an OOD detection statistic: samples with $\Delta \ell_c > \tau$ are classified as OOD, where $\tau$ is chosen to achieve a $95\%$ true positive rate on a held-out ID validation set.

\section{Experiments and Results}
\textbf{Datasets and Setup:}
We evaluate our method on three medical imaging benchmarks: Kvasir-v2 \cite{Pogorelov:2017:KMI:3083187.3083212}, NCT-CRC-HE-100K and CRC-VAL-HE-7K (collectively referred to as histopathology) \cite{kather2018100}, and the Retinal Optical Coherence Tomography (OCT) Dataset \cite{kermany2018identifying}. 
Following \cite{chhetri2025nero}, for Kvasir-v2, we treat three healthy landmarks (Normal Cecum, Normal Pylorus, Normal Z-line) as ID and the remaining five classes as OOD.
For histopathology, cancer-associated stroma (STR) and colorectal adenocarcinoma epithelium (TUM) are defined as OOD, with the remaining classes as ID.
For OCT, we adopt a class-level split: Normal and CNV are ID, while DME and DRUSEN are held out as OOD, evaluating the detection of anatomically unseen retinal pathologies.
All images were resized to 224x224.\\

\noindent\textbf{Implementation Details:}
We conduct our experiments on ViTs, as they have achieved SOTA performance in various medical benchmarks and they operate in dense, high-dimensional semantic spaces that are often harder to interpret.
The model backbone for Kvasir-v2, Histopathology dataset, and the Retinal OCT dataset is initialized with pretrained weights from GastroNet-5M \cite{jong2025gastronet}(ViT-S/16), PathoDuet \cite{HUA2024103289}(ViT-B/16), and ImageNet pretrained weights (ViT-S/16), respectively.
To account for the low volume of training data in Kvasir-v2 (2,400 compared to 63,520
for the Retinal OCT dataset and 75,237 for the Histopathology dataset), which might limit the quality of representations learned by SAE, we pretrain the SAE for Kvasir-v2 on GastroNet-5M\cite{jong2025gastronet}.
We employ the ViT-Prisma framework \cite{joseph2025prisma} to train SAEs.
Full implementation and training details will be provided in the released code.

\noindent\textbf{Ours vs. Baselines:} Table~\ref{results} compares our method against OOD detection baselines on Kvasir-v2, OCT, and Histopathology using AUROC and FPR95 metrics.
Our approach achieves state-of-the-art performance on OCT and Histopathology, reaching 93.67\% and 95.28\% AUROC with the lowest FPR95 (24.8\% and 23.17\%, respectively).
The moderate performance on Kvasir-v2 likely arises from the limited quality of learned semantic concepts.
Despite large-scale endoscopic pretraining, the downstream dataset comprises only 800 samples per class, thereby constraining the SAE’s capacity to extract highly discriminative, class-specific directions, given its unsupervised training paradigm.
In contrast, OCT and Histopathology contain substantially larger training sets (>50k samples), enabling richer and more stable concept discovery. 
\begin{table}[!t]
\caption{Metrics include AUROC (higher is better) and FPR95\% (lower is better), both reported as percentages. The best results are highlighted
in bold, while the second-best and third-best results are underlined.}
\centering
\begin{tabular}{l c c c c c c c c c c }
\hline
& & &\multicolumn{2}{c}{Kvasir-v2} & \text{      }& \multicolumn{2}{c}{OCT} & \text{      }& \multicolumn{2}{c}{Histopathology}\\
Method & Venue &  & AUC & FPR95 &  & AUC  & FPR95 &  &  AUC & FPR95 \\
\hline
Energy  & NeurIPS’20 & & 63.64 & 67.90 & & 83.84 & 38.08 & & 93.41 & \underline{23.74}  \\
Entropy  & IJCNN’21 & & 50.99 & 97.20 & & 51.81 & 96.38 & &92.25 & 32.14 \\
MSP   &  ICLR’17  & & 50.99 & 97.20 & &51.81 & 96.38 & &92.23 & 32.36 \\
Mahalanobis &  NeurIPS'18 & & \textbf{84.10} & \textbf{40.68} & &\underline{90.24} & \underline{28.45} & &93.53 & 26.44 \\
Vim     & CVPR’22 & & \underline{82.64} & \underline{40.74} & &\underline{89.79} & \underline{28.72} & &\underline{93.56} & 25.67 \\
NECO     & ICLR’24 & & 59.96 & 71.86 & &84.81 & 36.38 & &\underline{94.67} & 23.84 \\
MaxLogit   & ICML’22  & & 63.64 & 67.90 & &83.84 & 38.08 & &93.37 & 23.81\\
ODIN    & ICLR’18 & & 63.13 & 75.18 & &82.98 & 39.01 & &91.17 & 27.53 \\
Energy+React & NeurIPS'21 & & 63.62 & 67.64 && 84.90 & 36.33 && 93.41 & \underline{23.74} \\
NERO    & MICCAI’25 & & 76.58 & \underline{55.32} && 76.77 & 49.59 && 74.35 & 83.2 \\
GradNorm  & NeurIPS’21  & & 54.47 & 78.62 && 82.92 & 42.91 && 88.78 & 33.20 \\
I-ODIN   &  CVPR'26 & & 63.39 & 76.68 && 83.55 & 38.52 &&91.56  & 27.57  \\
\textbf{CAPS (Ours)} & - & & \underline{79.58} & 55.38 && \textbf{93.67} & \textbf{24.80} && \textbf{95.28} & \textbf{23.17} \\
\hline
\end{tabular}
\label{results}
\end{table}

\begin{figure}[!htb]
    \centering
    \includegraphics[trim= 0cm  0cm 0cm 0cm, width=0.975\textwidth]{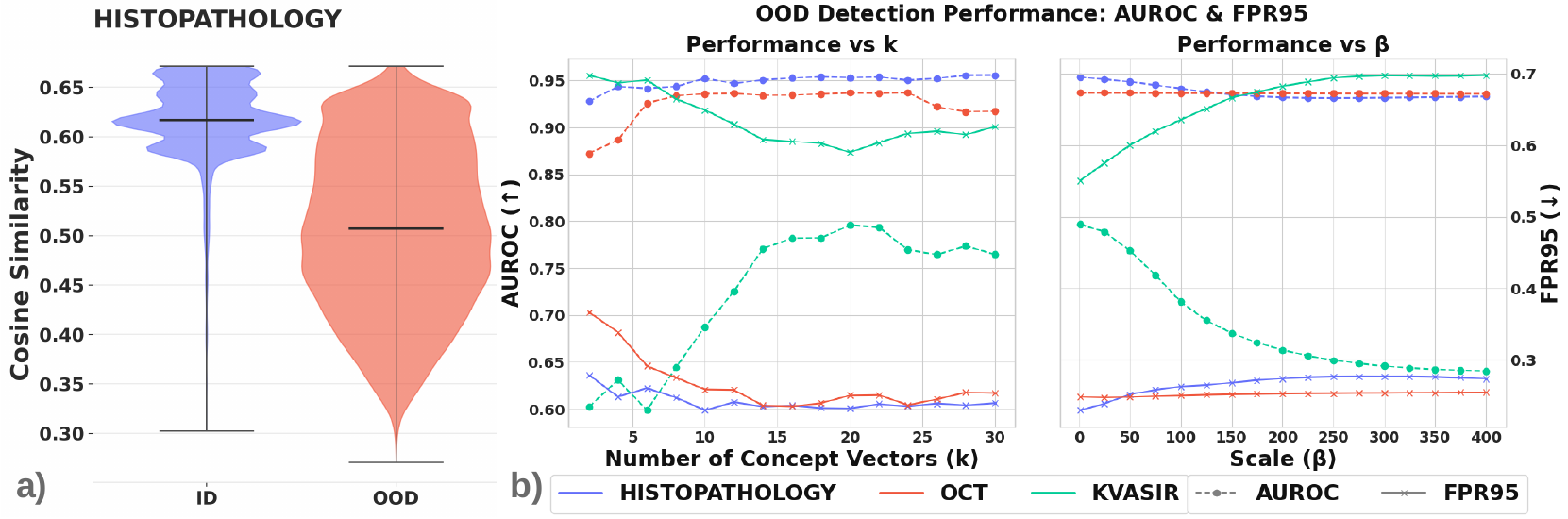} 
    \caption{(a) Cosine similarity between the class concept vector $v_c$ and sample activations for ID and OOD data in the histopathology dataset (similar trends are observed for Kvasir and OCT). (b) Robustness analysis with respect to the number of concept vectors $k$ and the scaling parameter $\beta$ across Histopathology, OCT, and Kvasir datasets.}
    \label{fig:robustness}
\end{figure}
\noindent\textbf{The effect of hyperparameter: } In Fig.~\ref{fig:robustness}(b), we evaluate the robustness of our method with respect to the hyperparameters $k$ and $\beta$, which denote the number of concepts per class and the perturbation strength, respectively.
The results show that performance remains largely stable across a wide range for the histopathology and OCT datasets.
In contrast, on Kvasir-v2, we observe noticeable instability for both values $k$ and $\beta$.
We hypothesize that this behavior is linked to the quality of the learned semantic concepts, as discussed  Section~\ref{Conclusion}.\\
\subsection{Qualitative Interpretation of Learned SAE Concepts:}
\label{qualitativeSection}
\begin{figure}[!htb]
    \centering
    \includegraphics[trim= 0cm  0cm 0cm 0cm, width=1\textwidth]{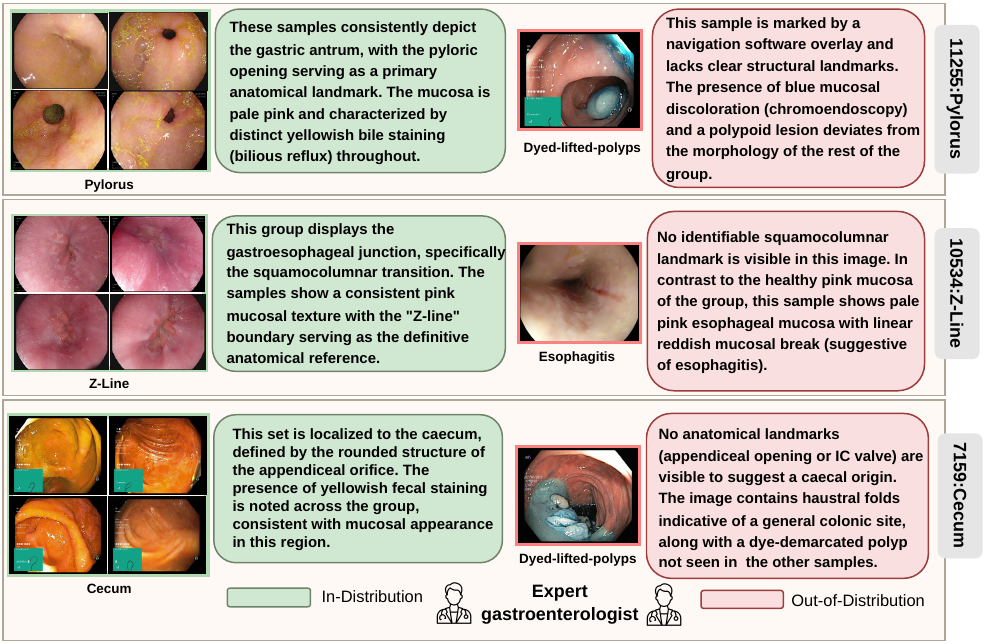} 
    \caption{
Clinical Grounding and OOD Contrast of Sparse Concept Latents.
Each row pairs a base-model predicted class with an SAE latent (concept). Left (green): ID clusters showing the four images with the highest latent activations.
Right (red): An OOD sample with low activation.
While all images in a row share the same predicted class, green and red boxes provide expert clinical descriptions for the ID cluster and OOD outlier, respectively.
Labels beneath denote the ground-truth original class.
       }
    \label{fig:qualitative}
\end{figure}
\noindent To interpret the sparse concepts, we ground each SAE feature in real-image exemplars to assess whether it corresponds to a coherent, interpretable visual pattern. 
Unlike attribution or network dissection methods \cite{selvaraju2017grad,bau2017network}, SAEs learn explicit directions in activation space, enabling direct intervention and representation-level analysis \cite{stevens2025interpretabletestablevisionfeatures}. 
For each class, we identified the top-$k$ defining concepts (Sec.~\ref{method:collectingConcept}, $k{=}20$) and clustered their top-20 activating validation images.
We extended this framework to characterize base-model failure modes on OOD data by introducing contrastive outliers: samples incorrectly assigned to a class with high confidence despite negligible SAE concept activation.
A blinded clinical expert was tasked with: (i) identifying shared semantic patterns within clusters and (ii) isolating substantive outliers.
The expert consistently identified shared clinical markers in ID clusters, including anatomical landmarks (pyloric opening, appendiceal orifice) and physiological features (bilious reflux or fecal staining).
Conversely, the expert isolated the OOD sample, noting an absence of these class-dominant features, a finding that correlates with the low activation magnitudes recorded for those latents (Fig. \ref{fig:qualitative}).
This experimental design qualitatively demonstrates that while highly activating samples exhibit consistent, human-recognizable features, the contrastive results reveal that the base model may assign high-confidence labels to OOD samples, but it does so in the absence of the expert-verified clinical features that drive the ID predictions.
By surfacing this concept nullity in OOD samples, we provide a mechanistic justification for our detection framework: it effectively identifies instances where the model's prediction is disconnected from the interpretable feature space learned by the SAE.

\section{Conclusion}\label{Conclusion}
We introduced a class-conditioned activation perturbations (CAPS) method in representation space for OOD detection, where concept vectors are derived from SAEs.
Validated across endoscopy, histopathology, and ophthalmic imaging, the approach demonstrates cross-domain generalizability. 
Clinician evaluation confirmed that SAE latents correspond to clinically meaningful visual patterns, validating the interpretability of the learned sparse representation space.
Importantly, we showed that this interpretability extends to OOD detection: OOD samples lack class-dominant concept activations, a deficiency verified quantitatively via low activation magnitudes and qualitatively through expert review.
In low-data regimes, the difficulty of capturing semantically coherent directions may impact reliability.
While modality-level pretraining reduces these effects, developing more robust SAE training strategies for data-constrained healthcare environments remains an important direction for future work.
We view this work as a step toward unifying representation-level interpretability with reliability assessment in deep neural networks and hope encourages further research.
\bibliographystyle{splncs04}
\bibliography{bib}
\end{document}